%% file: ecai-sample-and-instructions.tex
\documentclass{ecai} 

\usepackage{latexsym}
\usepackage{amssymb}
\usepackage{amsmath}
\usepackage{amsthm}
\usepackage{booktabs}
\usepackage{enumitem}
\usepackage{graphicx}
\usepackage{color}
\usepackage{ulem}
\usepackage{multicol, multirow}
\usepackage{ragged2e}
\usepackage{caption}
\usepackage{float}
\newcommand{\BibTeX}{B\kern-.05em{\sc i\kern-.025em b}\kern-.08em\TeX}

\definecolor{brightlavender}{rgb}{0.75, 0.58, 0.89}

\begin{document}


\begin{frontmatter}


\paperid{123} 


\title{Reinforcement Learning for Efficient Returns Management}


\author[A]{\fnms{Pascal}~\snm{Linden}\thanks{Corresponding Author. Email: pascal.nicolai.linden@iais.fraunhofer.de}}
\author[A]{\fnms{Nathalie}~\snm{Paul}}
\author[A]{\fnms{Tim}~\snm{Wirtz}}
\author[A,B]{\fnms{Stefan}~\snm{Wrobel}} 

\address[A]{Fraunhofer Institute for Intelligent Analysis and Information Systems IAIS}
\address[B]{University of Bonn}


\begin{abstract} 
In retail warehouses, returned products are typically placed in an intermediate storage until a decision regarding further shipment to stores is made.
The longer products are held in storage, the higher the inefficiency and costs of the returns management process, since enough storage area has to be provided and maintained while the products are not placed for sale. 
To reduce the average product storage time, we consider an alternative solution where re-allocation decisions for products can be 
made instantly upon their arrival in the warehouse allowing only a limited number of products to still be stored simultaneously. We transfer the problem to an online multiple knapsack problem and propose a novel reinforcement learning approach to pack the items (products) into the knapsacks (stores) such that the overall value (expected revenue) is maximized. Empirical evaluations on simulated data demonstrate that, compared to the usual offline decision procedure, our approach comes with a performance gap of only $3\%$ while significantly reducing the average storage time of a product by $96\%$.
\end{abstract}

\end{frontmatter}


\section{Introduction}
Managing returns is a central process in the retail supply chain as it has a high impact on the companies' costs and their sustainability \citep{rogers2002returns}. Artificial Intelligence and Machine Learning particularly have shown great potential in optimizing returns management \citep{ambilkar2022product} by, e.g., forecasting product returns \citep{cui2020predicting}, identifying causes for returns \citep{cheng2024customers} or optimizing the logistics for successfully utilizing the given transport capacities \citep{toth2014vehicle}.
However, the process of decision-making in the warehouses to route the products internally was hardly addressed yet.

Here, we consider the return allocation decision in the warehouse whereby products are re-assigned to stores for sale. 
Common practice is to intermediately store returned products for a few days in the warehouse and then allocate the batch of products (offline approach). It allows the retailer to make a more globally optimized re-allocation decision given the knowledge about the batch of products.
On the other hand, the high product storage times lead to an inefficient and costly decision-making process.
In particular, the costly storage space is exclusively used for making proper allocation decisions. Products have to be handled repetitively in terms of scanning and moving within the warehouse. In addition, selling of products is delayed.
Ideally these issues could be addressed by an online allocation approach where allocation decisions are made on-the-fly upon a product's arrival in the warehouse. However, the quality of the allocation decisions would be affected as the relevant information about future incoming products cannot be taken into account at all. We want to tackle this challenge and propose a relaxed online allocation approach, that still produces high-quality allocation decisions while the product storage time is significantly reduced.

In particular, we propose a novel reinforcement learning (RL) approach called \textit{PostAlloc} which allows only a limited number of products to be stored at the same time.
RL has become powerful for learning heuristics for complex combinatorial optimization problems like routing \citep{kool2018attention}, scheduling \citep{chen2019learning} and also knapsack problems \citep{sur2022deep} which can be viewed as allocation problems: Items (products) have to be packed into knapsacks (stores) such that the total value (expected revenue) is maximized while respecting the knapsacks' capacity constraints (stores' capacities). 
Our approach builds on the RL model proposed by \citet{kong2018new}, which implements a neural network-based decision policy to solve an online \textit{single} knapsack problem. We here extend their approach to the case of multiple knapsacks and, in addition, adapt it such that allocation decisions of products are allowed to be postponed by intermediately storing the product. 
To evaluate the impact of our method on the quality of allocation decisions as well as the efficiency of the returns management process, the performance of our approach is evaluated from two angles. First, we report the achieved solution quality in comparison to an offline mathematical solver which can compute optimal allocation solutions. Second, we compare our resulting average product storage time to the one of the common practice offline decision procedure.
\newline\newline
In the following, we specify the considered return allocation problem and translate it to the online multiple knapsack problem. Section \ref{sec_relwork} provides an overview of the related work. In Section \ref{sec_methods} we discuss our solution approach \textit{PostAlloc}. We describe the design of the considered allocation problem as a RL problem as well as its implementation. Empirical evaluations of our approach are performed on simulated data in Section \ref{sec_eval}. We consider various data distributions and benchmark the results of our approach. 
Section \ref{sec_concl} summarizes the results and outlines plans for future work.

\section{Problem Statement}
\label{sec_problemStatement}

As a first step to assess the performance of RL for return allocation decisions given a limited-size storage, we present a simplified problem formulation where further requirements can be integrated in the future.
It is motivated and described in Section \ref{subsec_alloc} and mapped to the online multiple knapsack problem in Section \ref{subsec_knap}.

\subsection{Motivating Example}
\label{subsec_alloc}
Consider a retail company consisting of multiple stores and a central warehouse. The company's goal is to maximize its profit by ideally supplying its stores with the returns arriving at the warehouse. In this example, we can describe the entities involved in the allocation process as follows: There are $K$ stores to which returns can be delivered. Depending on the size of the store, it can accept a certain quantity of returns in addition to the current stock. For the sake of simplicity, we can represent a store's available space for returns as its capacity $C$. Each product allocated to a store consumes a portion of the remaining space. We call the amount of space a product consumes $w$. At the same time, a product also has a value $v$ for the company indicating the product's lucrativeness if it is placed in stores for sale. E.g., it could be given by the price of the product in relation to the time it takes to sell the product.

Now, for each returned product arriving in the warehouse, an allocation decision has to be made. That means deciding to which of the $K$ stores a product should be delivered. 
If none of the physical stores has enough capacity for a product or if their capacity should be saved for other possibly more lucrative products, we assume a product to be shipped to a different storage unit containing the products for online sales. 

In our envisioned returns management procedure, the decision of assigning an incoming product to any of the physical stores or the online store can be instantly made upon the product's arrival in the warehouse or postponed by placing it in a limited-size intermediate storage area.

\subsection{Translation to the Knapsack Problem}
\label{subsec_knap}
The above described allocation problem can be translated to an online multiple knapsack problem where $K$ knapsacks (stores) with individual capacities need to be optimally packed with items (products), which are characterized by their value $v$ and weight $w$. The items arrive one by one as part of a sequence of size $N$. The goal is to maximize the total sum of values in the knapsacks $V$ while the sum of the items' weights $W_j$ in each knapsack $j$ must not exceed its capacity $C_j$.

In its classic formulation, there is the possibility to \textit{accept} an item for being placed in a specific knapsack (physical store) or to \textit{reject} it (online store). To model the envisioned procedure, we introduce a new additional action called \textit{postpone} to place the item in a limited-size buffer (intermediate storage area) and delay the decision until the end of the item sequence.


In the following, we stay with the terminology of the knapsack problem.

\section{Related Work} 
\label{sec_relwork}
To tackle the classic formulation of the NP-hard knapsack problem \citep{karp2010reducibility}, various hand-crafted heuristics have been designed which typically make use of the items' value-to-weight ratios. In an offline setting where all items are observed at once, the heuristics rely on sorting the items regarding these ratios and achieve high performance results with that \citep{pisinger2005hard}. In the much more challenging online setting, one can only (dynamically) define thresholds on the value-to-weight ratios and the performance of such online heuristics highly depends on the data \citep{chakrabarty2008online}. To allow performance improvements, relaxations on the online setting have been discussed like the introduction of a resource buffer with higher capacity than a knapsack which allows to pre-collect a subset of the incoming items and subsequently solve the problem offline \citep{han2019online}. Reinforcement Learning (RL) has shown to be able to learn heuristics which outperform hand-crafted heuristics for various problems like routing \citep{kool2018attention}, scheduling \citep{chen2019learning} and also the knapsack problem. For the latter, the focus so far has been on offline settings \citep{afshar2020state,sur2022deep} or on online settings with only a single knapsack \citep{kong2018new, tu2023deep}. \citet{kong2018new} use a simple neural network to learn a policy which manages to imitate a threshold-based online algorithm known to behave close to optimally on the considered dataset. As inputs, their network receives the value and weight of the current item, its relative position in the sequence, and the fraction of the capacity that is occupied by the accepted items so far. If an accepted item fits into the knapsack, the reward is defined as the item's value, otherwise it is set to zero. Their analysis of the model behavior indicates that the items' value-to-weight ratios are of key importance for the action decisions.
We build on their approach and extend it to solve an online \textit{multiple} knapsack problem with the additional opportunity to postpone the decision about some items to the end of the sequence (\textit{PostAlloc}). For a detailed description of the model differences, see Appendix \ref{diff_Kong}. To the best of our knowledge, we present the first RL approach for solving an (relaxed) online multiple knapsack problem.

\section{Methods}
\label{sec_methods}

In the following, we describe our RL-based approach \textit{PostAlloc} to solve the online multiple knapsack problem stated in \ref{subsec_knap}. More precisely, we present the design as a RL problem by defining states, actions and rewards in Section \ref{sec_RL}. The neural network-based model architecture is then explained in Section \ref{sec_models}.

\subsection{Reinforcement Learning Formalization}
\label{sec_RL}
\paragraph{State} The state at time step $i$ is defined as $s_i = (v_i ,w_i , \frac{i}{N}, \frac{B_i}{B},\{\frac{W_{i,j}}{C_j}\}_{j\in [K]})$, providing information about the current item's value and weight, the item’s relative position within the sequence, the fraction of occupied space in the buffer $B$ and for each knapsack the fraction of the capacity that is already occupied by accepted items. The sequence of fractions $\{\frac{W_{i,j}}{C_j}\}_{j\in [K]}$ is sorted in ascending order at each time step $i$, i.e., the knapsack with the largest remaining capacity always comes first.

\paragraph{Action} Given a state $s_i$, the agent selects an action $a_i$ from the action space $A=\{$\textit{accept}, \textit{reject}, \textit{postpone}$\}$ which consists of the two basic actions \textit{accept} and \textit{reject}, as well as an additional action we call \textit{postpone}. The basic actions translate to placing an item in the knapsack which has the largest remaining capacity, or not packing the item at all.
The newly introduced action \textit{postpone} allows the agent to store the current item in an intermediate storage called buffer. This defers the final decision to \textit{accept} or \textit{reject} the item until the end of the sequence. Yet, if the buffer is already full at the time of a postpone decision, the item with the worst value-to-weight ratio is removed from the buffer and an accept or reject decision must be made.
We note that an accepted item is only indeed packed if the knapsack's capacity is not exceeded and is rejected by the environment otherwise.

In this model, an item cannot be packed into a specifically chosen knapsack $j$. The current item can only be placed into the knapsack with the largest remaining capacity similar to \citet{sur2022deep}. Otherwise, the item must be postponed or rejected.
This allocation heuristic enforces the capacities across all knapsacks to increase more evenly. This also seems desirable for the returns management application as it avoids cases in which many products are delivered to only one or a few stores while other stores are hardly re-stocked. Moreover, the chance for a (valuable) item to still fit into a knapsack is always highest for the one with the largest remaining capacity.

\paragraph{Reward} If the item at step $i$ is accepted and fits into the knapsack with the largest remaining capacity, the reward is given by the item's value $r_{i+1} = v_i$. While the reward for rejecting an item is $0$, choosing the \textit{postpone} action receives a fixed negative reward of $-1$. That is because this action causes an increase in the intermediate storage time, and, therefore, it should not come for free.

\subsection{Model Overview}
\label{sec_models}
\begin{figure}[ht]
\centering
\includegraphics[width=8.5cm]{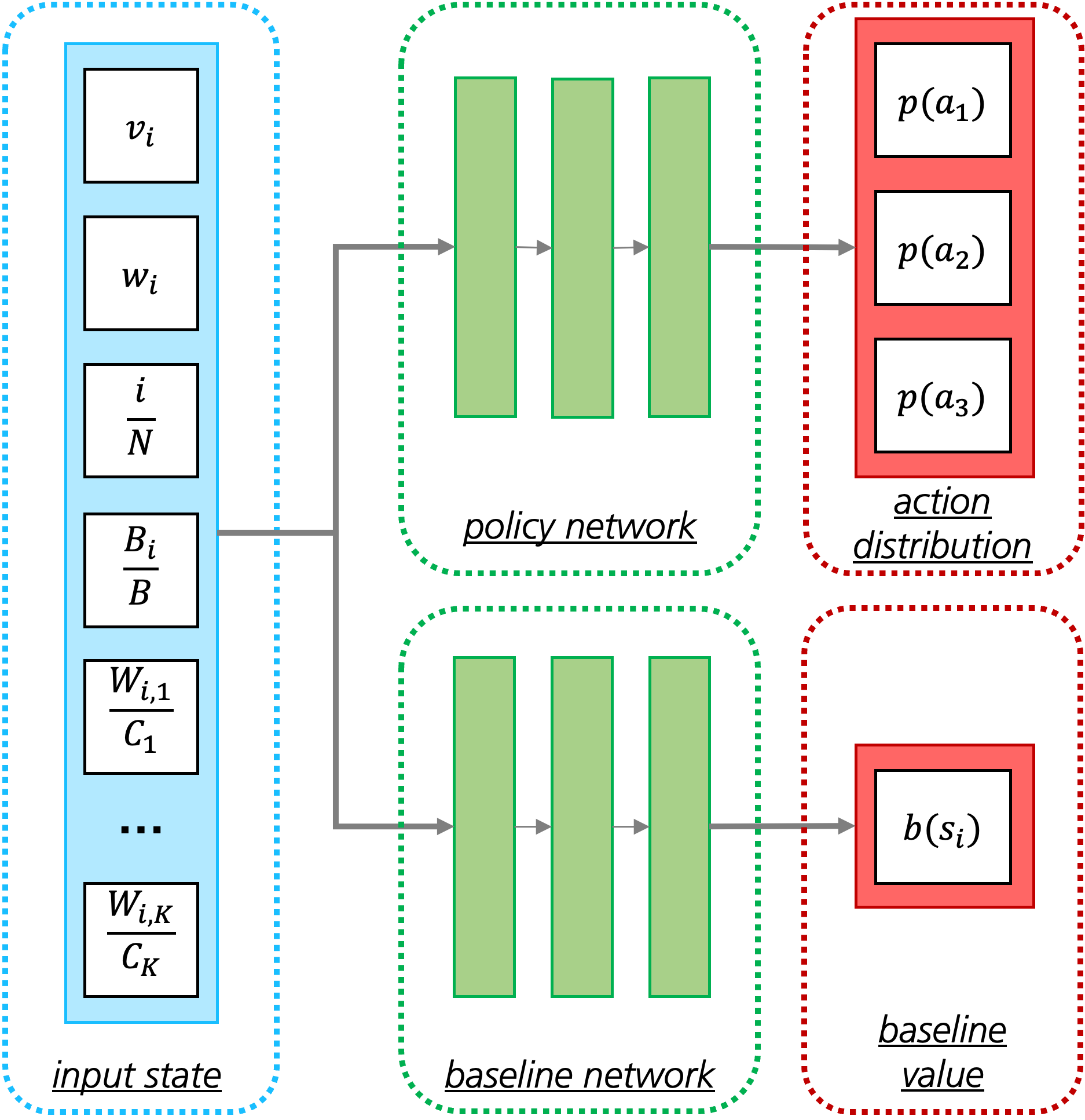}
\caption{Model architecture of \textit{PostAlloc}. At each step $i$, the input state $s_i$ is fed into the policy and the baseline network. The policy network outputs a probability distribution over the actions \textit{accept}, \textit{reject}, and \textit{postpone} while the output of the baseline network is used in both networks' gradient-based parameter updates.}
\label{fig:architecture}
\end{figure}

\paragraph{Policy} The decision policy is implemented as a fully connected feed-forward neural network, cf. Figure \ref{fig:architecture}. As input, the network receives the current state $s_i$ at step $i$ as defined above. After the three hidden layers with 50 neurons each, the network creates a three-dimensional output vector which is normalized by a softmax function to produce a probability distribution over the three actions in the action space. The network is trained using the REINFORCE algorithm with a baseline \citep{williams1992simple}.

\paragraph{Baseline network} We use a second neural network that learns to provide a baseline value for the REINFORCE algorithm depending on the current state. Except for the last layer, the baseline network consists of the same structure as the policy network, cf. Figure \ref{fig:architecture}. The last layer of the baseline network reduces the dimension of the output to a single value representing the baseline value of the input state. We train this network simultaneously to the policy network using a gradient ascent method.

\section{Empirical Evaluation}

We empirically evaluate and benchmark our approach \textit{PostAlloc} on various simulated data setups.
The procedure for generating different problem setups is described in Section \ref{subsec_datagen}. Section \ref{subsec_bench} gives an overview of the considered benchmarks which range from weak online heuristics to strong offline solvers. The considered evaluation metrics are defined in Section \ref{subsec_metrics}. At last, we discuss the experimental results in
Section \ref{subsec_exp}.

\label{sec_eval}

\subsection{Data generation}
\label{subsec_datagen}
For our experiments, we use twelve data setups which differ in the total number of knapsacks and the amount of correlation between the items' weights and values. More precisely, we examine settings with $K = [1, 3, 5, 7]$ knapsacks, as well as with no, weak, or strong correlation between the items' weights and values following \citet{martello1990knapsack}. For each of these combinations, we generate $2500$ sequences consisting of $N=200$ items. The capacities of the knapsacks are calculated such that they add up to approximately $50\%$ of each sequence's sum of weights. For more details on the data generation, see \ref{data_gen}.

\subsection{Benchmarks}
\label{subsec_bench}
To evaluate our approach \textit{PostAlloc}, we compare it with several other ranging from low- to high-performing solution methods. Methods designed for one knapsack are used for multiple knapsacks by considering only the one knapsack with the highest remaining capacity in the respective decision step, in analogy to our approach.\newline
First, we measure the performance of a \textit{random} agent whose policy is to flip an unbiased coin to decide whether to accept or reject an item. A second straightforward solution is an agent that always accepts an incoming item as long as it fits into the current knapsack (\textit{Take all}). We additionally consider two problem-specific heuristics. 
The first is given by the online \textit{CZL} algorithm \citep{chakrabarty2008online} which accepts an item if its value-to-weight-ratio is larger than a dynamically increasing threshold.
The second is an offline greedy algorithm \citep{martello1990knapsack} (\textit{GreedyOff}) which, knowing all items of the sequence, iteratively packs the item with the current highest value-to-weight ratio while respecting the capacity constraints.
We also consider the approach proposed by \citet{kong2018new} as our model is built on top of their ideas. Finally, we use Google’s library OR-Tools \citep{ortools} to solve each sequence optimally in an offline setting.

\subsection{Evaluation Metrics}
\label{subsec_metrics}
We evaluate our approach from two perspectives by looking at the quality of the generated optimized solutions as well as the impact on the efficiency of the returns management process. The optimization performance is evaluated in terms of the relative optimality gap compared to OR-Tools as $\frac{V_{\text{or}}-V_{\text{alg}}}{V_{\text{or}}}$, where $V_{\text{x}}$ with $\text{x} \in \{\text{or},\text{alg}\}$ denotes the total sum of values of the packed items for OR-Tools and any other considered algorithm.
The efficiency impact is evaluated in terms of the reduction in the average storage time of an item (product) compared to the typical offline scenario. To compute the product storage times, let $I = [1, 2, 3, ..., N]$ denote the discretized time window covering the arrival of product $i$ at time step $i$ for $i \in [N]$, with respect to a certain unit of time. In the offline scenario, the decision regarding the allocation of a product is made after the arrival of all products at time step $N+1$. Product $i$ is thus stored for time $\delta^B_i = N+1-i$, leading to the average product storage time of $\overline{\delta^B} = \frac{\sum_{i=1}^N (N+1-i)}{N} = \frac{1}{2}(N+1)$. In our approach, product $i$ is stored for time 

$$\delta_i = 
\begin{cases} 
\text{number of time steps in buffer}, & \text{if the decision is postponed,} \\ 0, & \text{otherwise.}\end{cases}$$

\noindent The corresponding average product storage time is given by \newline$\overline{\delta} = \frac{\sum_{i=1}^N \delta_i}{N}$.

\subsection{Experiments}
\label{subsec_exp}

\paragraph{Hyperparameters} For training each of the networks, we use the Adam optimizer \citep{kingma2014adam} with a learning rate of $10^{-4}$ following \citet{kong2018new}.
For the newly introduced buffer, a fixed size of $|B| = 0.05 \cdot N$ was chosen. With the $N=200$ items per sequence, this translates to a total number of $10$ items that can be stored in the buffer at the same time. The number of training epochs was set to $100$ epochs.

\paragraph{Results}

\begin{figure}[h!]
\centering
\includegraphics[width=9cm]{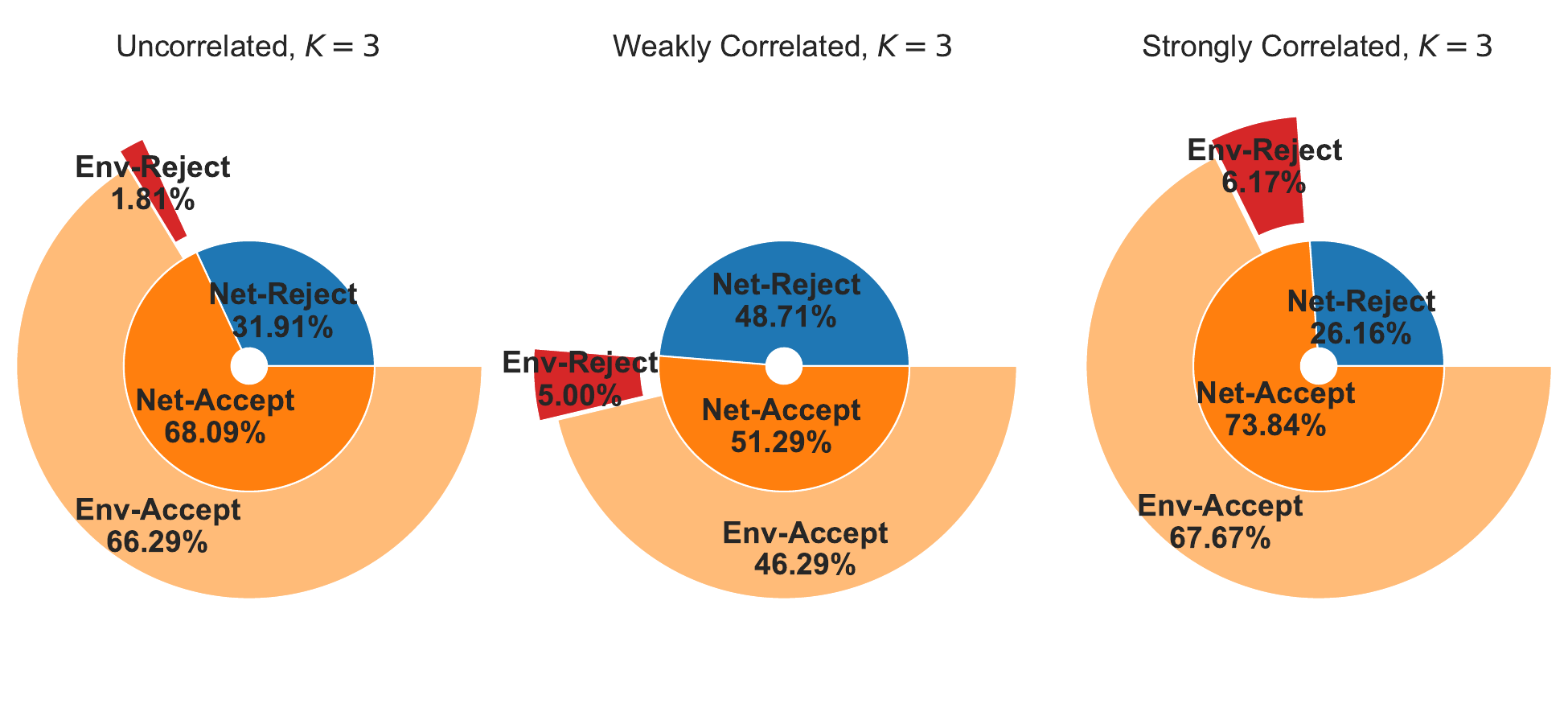}
\includegraphics[width=9cm]{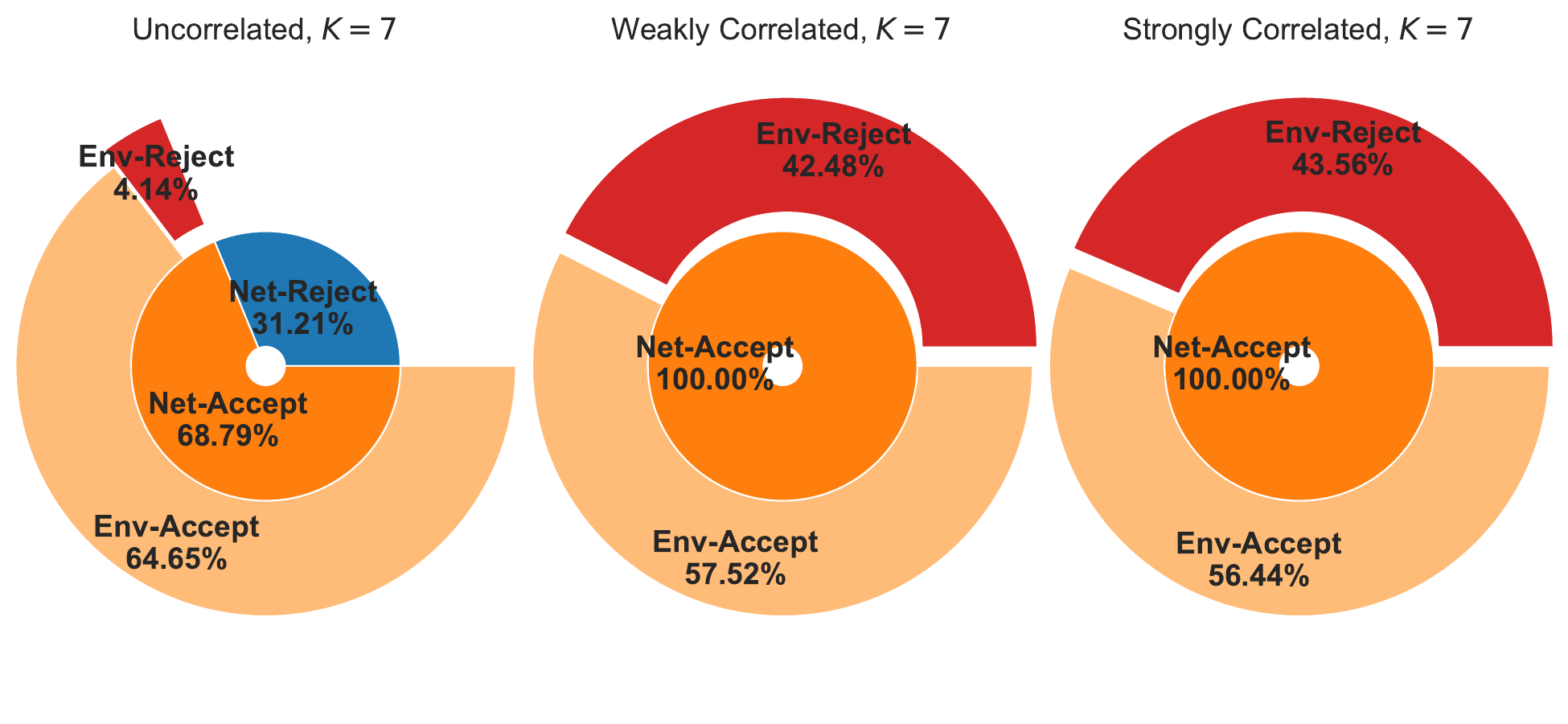}
\vspace{-1cm}
\caption{Final \textit{accept} and \textit{reject} decisions of the network on all datasets with $K=3$ (top) and $K=7$ (bottom). Every item must be accepted or rejected independent of whether the decision has been postponed previously. The inner circle depicts the network decisions, while the \textit{accept} decisions of the network are refined on the outer layer into the decisions of the environment to accept or reject the items. The environment always enforces the capacity constraints to be fulfilled. Thus, all cases marked with \textit{Env-Reject} are attempts of the network to accept an item even though it does not fit into the current knapsack anymore.
For $K=3$ (top), a well-balanced distribution of \textit{accept} and \textit{reject} decision has been learned for all datasets. For $K=7$ (bottom) this is only the case for the uncorrelated dataset. For those with correlation, the networks' policy converged into a take-all strategy meaning that no useful behavior was learned.}
\label{fig:decision-k3k7}
\end{figure}

Tables \ref{tab:results_Uncorr}, \ref{tab:results_weakCorr} and \ref{tab:results_strongCorr} show that our approach \textit{PostAlloc} generates high-quality solutions comparable to OR-Tools for most of the problem setups. The only exceptions are the weakly and strongly correlated cases with $K=7$. Here, the bottom middle and bottom right plot of Figure \ref{fig:decision-k3k7} visualize that the resulting decision policy is identical to the take-all strategy.
Since the state dimension grows with the number of knapsacks, the result indicates that the scalability of our approach needs to be improved.
For all other cases, our approach significantly outperforms the online algorithms (Random, Take all, CZL, Kong) and gets close to the offline greedy algorithm, even outperforming it in the strongly correlated cases $K=3$ and $K=5$, cf. Table \ref{tab:results_strongCorr}. This is exemplified by the top line plots in Figure \ref{fig:decision-k3k7}, showing that for $K=3$, the learned policies have a well-balanced distribution of accept and reject actions where only in very few cases the capacity constrains had to be enforced by the environment.

The performance improvement compared to the RL model of Kong demonstrates the benefit of introducing a buffer allowing to postpone decisions.
Interestingly, during inference our approach makes only use of postpone actions for the uncorrelated dataset.
This can be explained by the fact that in the uncorrelated setting there is more variation between the weights and values. We overall observe that the majority of items that are rejected or accepted instantly, have either very bad or good value-to-weight ratios. The value-to-weight ratios of postponed items are rather average and, therefore, it seems reasonable to defer the final decision for those. In the datasets with more correlation, the value-to-weight ratios are more similar. Accordingly, there is less benefit from a delayed decision as this decision comes with a negative impact on the overall reward and it makes sense that the models focus on maximizing the gain through the basic actions.
Summarizing the optimization perspective, the performance of our \textit{PostAlloc} model is only $2.88\%$ below that of OR-Tools even though it does not require the knowledge about the full sequence in advance.

From the efficiency perspective, we reduce the average product storage time by $100\%$ on the weakly and strongly correlated datasets compared to the offline scenario, since in these cases the buffer is not made use of as described above. On the uncorrelated datasets where postpone actions have been used, the reduction is still tremendously high with $96.1\%$ on average.
For an extended discussion of the performance results, see \ref{app:results}.

\begin{table}[h!]
\caption{Results on the uncorrelated data sets in terms of the relative gap to OR-Tools. The best result of the five online approaches is highlighted.}
\centering
\begin{tabular}{c|cccc}
    \hline
     & \multicolumn{4}{c}{Uncorrelated} \\
     & K=1 & K=3 & K=5 & K=7 \\ \hline 
Random  &  40.08\% & 44.49\% & 39.84\% & 38.92\% \\
Take all &  36.76\% & 38.41\% & 32.17\% & 29.29\%  \\
CZL  &   26.17\% & 32.20\% & 27.91\% & 34.41\%  \\
Kong  &   2.31\% & 2.77\% & 2.65\% & 4.24\%  \\
PostAlloc  &   \textbf{1.20}\% & \textbf{1.59}\% & \textbf{2.47}\% & \textbf{2.42}\% \\ \hline
GreedyOff  &  0.04\% & 0.30\% & 0.52\% & 1.20\%  \\
\hline
\end{tabular}
\label{tab:results_Uncorr}
\end{table}

\begin{table}[h!]
\caption{Results on the weakly correlated data sets in terms of the relative gap to OR-Tools. The best result of the five online approaches is highlighted.}
\centering
\begin{tabular}{c|cccc}
    \hline
     & \multicolumn{4}{c}{Weakly correlated} \\
     & K=1 & K=3 & K=5 & K=7 \\ \hline 
Random  &  18.37\% & 14.89\% & 18.61\% & 17.31\% \\
Take all &  15.30\% & 11.57\% & 13.62\% & 12.62\% \\
CZL  &   21.92\% & 17.69\% & 38.85\% & 15.26\%  \\
Kong  &   12.38\% & 11.57\% & 13.55\% & \textbf{12.50}\%  \\
PostAlloc  &   \textbf{0.99}\% & \textbf{1.85}\% & \textbf{3.23}\% & 12.62\% \\ \hline
GreedyOff  &  0.04\% & 0.22\% & 0.71\% & 0.82\%   \\
\hline
\end{tabular}
\label{tab:results_weakCorr}
\end{table}

\begin{table}[h!]
\caption{Results on the strongly correlated data sets in terms of the relative gap to OR-Tools. The best result of the five online approaches is highlighted.}
\centering
\begin{tabular}{c|cccc}
    \hline
     & \multicolumn{4}{c}{Strongly correlated} \\
     & K=1 & K=3 & K=5 & K=7 \\ \hline 
Random  &  8.08\% & 10.20\% & 10.16\% & 11.71\% \\
Take all &  4.50\% & 5.78\% & 5.47\% & \textbf{5.03}\%  \\
CZL  &   49.74\% & 36.33\% & 52.72\% & 39.98\%  \\
Kong  &   2.53\% & 5.78\% & 3.28\% & \textbf{5.03}\%  \\
PostAlloc  &  \textbf{0.65}\% & \textbf{0.96}\% & \textbf{1.61}\% & \textbf{5.03}\%  \\ \hline
GreedyOff  &  0.50\% & 1.71\% & 2.70\% & 4.48\% \\
\hline
\end{tabular}
\label{tab:results_strongCorr}
\end{table}

\section{Conclusion}
\label{sec_concl}

We proposed a novel reinforcement learning approach for returns management in retail
where products arriving in companies' warehouses have to be assigned to their future sales location. While incoming products are typically collected until an allocation decision is made (offline scenario), we allow only a limited number of products to be stored simultaneously to reduce the average product storage time. Experiments on simulated data for various problem setups show that our approach is still able to compete with the decision quality of the typical offline scenario, which has the privilege to compute optimal allocation decisions given the complete knowledge about all products. More precisely, our reinforcement learning approach reduces the average product storage time by 96\% with a performance gap of only 3\% compared to the offline scenario. Thus, it significantly increases the efficiency of the returns management process.
\newline\newline
In the future, we want to improve the performance of our approach for setups with a large number of possible sales locations. Therefore, we plan on extending the reward function to include penalties for violating store capacity constraints as well as not using capacities to the fullest. We also want to analyse the impact of the size of intermediate storage area on the performance of our approach.
Moreover, we want to further increase the practical relevance through a more complex and realistic modelling of the products' properties. E.g., a product's weight could be modeled with multiple dimensions precisely describing the product's geometric dimensions and weight. Allowing a product's value to depend on a store to represent, e.g., store-specific expected selling times is another interesting direction for future work.

\bibliography{mybibfile}

\input{appendix}

\end{document}

%% file: appendix.tex
\appendix
\section{Appendix}
\label{appendix}

\subsection{Differences to the Base Model}
\label{diff_Kong}
Our \textit{PostAlloc} model is based on the model proposed by \citet{kong2018new}. The three main differences are given in an extension of the state space, an extension of the action space, and a different version of the training algorithm.

\paragraph{State space extension} Their network architecture is designed for an online knapsack setting with a single knapsack. To utilize the same network for multiple knapsacks, there must be an additional mechanism to decide in which of the knapsacks an item is placed if the agent chooses to accept it. \citet{sur2022deep} present a heuristic solving this problem by pre-determining that an item may only be placed in the knapsack with the largest remaining capacity. Our framework implements this heuristic as well by extending the state space definition from Kong to include the sequence of sorted remaining knapsack capacities.

\paragraph{Action space extension} While the action space of the base model only consists of the two basic actions \textit{accept} and \textit{reject}, we introduce a new action called \textit{postpone}. This action comes with an additional artificial knapsack, the buffer. Whenever the \textit{postpone} action is chosen, the current item is added to the buffer to defer the decision to the end of the sequence. To provide the agent information about the buffer, we augment the state space further to include the fraction of occupied places within the buffer, similar to how a knapsack is represented in the state.

\paragraph{Training} \citet{kong2018new} mention using “the standard REINFORCE algorithm”. Instead, we train our network using a second neural network that learns to provide a baseline value depending on the current state as described in Section \ref{sec_models}.

\subsection{Data Generation}
\label{data_gen}

We have investigated settings with $K\in[1,3,5,7]$ knapsacks. Each knapsack $j$ requires an associated capacity $C_j$ to denote how much weight it can accommodate. We follow the ideas from \citet{martello1990knapsack} to generate \textit{similar capacities} for each knapsack. To calculate these, let $W_\sigma=\sum_{i=1}^Nw_i$ be the sum of all item weights from a sequence $\sigma$ of $N$ items. For the first $K-1$ knapsacks, their capacities $C_j$ are drawn from the interval $[(\beta-0.1)\cdot\frac{W_\sigma}{K}, (\beta+0.1)\cdot\frac{W_\sigma}{K}]$. For the $K$th knapsack, they assign a capacity of $c_K=\beta W_\sigma - \sum_{j=1}^{K-1}C_j$. Their choice of $\beta=0.5$ results in the knapsacks having a combined capacity of exactly half of the weight of all item weights from the sequence, i.e. $\sum_{j=1}^K C_j = 0.5 \sum_{i=1}^N w_i$. This method partitions the total $50\%$ of the sequence's weight similarly across the knapsacks. To introduce some variance in the capacity generation, we change the default value of $\beta$ and draw it uniformly from the interval $[0.47,0.53]$. This change is supposed to make it impossible for the network to learn an absolute dependency between the capacities and the sum of all weights in the sequence.
There are numerous options of how to randomly generate the weight $w_i$ and value $v_i$ of an item $i$. \citet{pisinger2005hard} gives a broad overview of $13$ different ways to generate instances. The three types of instances we have chosen are \textit{uncorrelated} instances, \textit{strongly correlated} instances, and \textit{weakly correlated} instances, following the choice from the broad investigation of \citet{martello1990knapsack}. In the uncorrelated case, weights and values are chosen uniformly at random from an interval $[1,R]$. Second, we use items that are strongly correlated, where only the weights are sampled uniformly. The values are defined based on the weights as  \(v_i = w_i + \frac{R}{10}\). Lastly, we modified the idea of weakly correlated data. In the definition from \citet{martello1990knapsack}, the values are sampled uniformly based on the sampled weights as \(v_i \sim U(w_i - \frac{R}{10}, w_i + \frac{R}{10})\). We have chosen to replace the uniform sampling with a sampling based on a normal distribution centered at the respective \(w_i\) with a standard deviation \(\sigma_{\mathcal{N}} = \frac{R}{10}\) as \(v_i \sim \mathcal{N}(w_i, \frac{R}{10})\). This keeps the idea of sampling most values from the interval as defined originally but, instead of the strict bounds, allows for a more smooth distribution of the resulting values. The items with weak or strong correlations are considered to be more challenging because the value-to-weight ratios do not vary as much as in the uncorrelated case. Therefore, given an item's value and weight, this ratio is not as good of an indicator whether to accept the item anymore. 
The choice of the range parameter $R$ varies depending on the literature. It starts with $R=10$ as the smallest range in \citet{sur2022deep}, increases to, for example, $R=100$ in \citet{tu2023deep} or \citet{yildiz2022reinforcement}, to $R=1000$ in \citet{martello1990knapsack}. In our setup, we have chosen $R=50$ and normalize the resulting weights and values in the state representation to a range of $[0,1]$.

Besides the general characteristics described above, we make some extra assumptions to represent a more realistic setting than a completely random dataset. The first assumption is that all sequences we generate for one dataset are based on the same set of items $I$ with $|I| = M < N$. This is supposed to model the situation where we have prior knowledge about which items can occur in future sequences, but we do not know which exact sequences to expect. We choose $M=50$ items per set $I$ in our experiments. Each item $i\in I$ consists of a triple $(w_i, v_i, \alpha_i)$ with $\alpha_i \in (5,20]$. The alpha value is for sampling the item's probability from a Dirichlet distribution. All alpha values are collected in a vector $\vec{\alpha} = [\alpha_1, ..., \alpha_M] \in (5,20]^M$. For every sequence $\sigma$, a probability distribution $P_\sigma \sim Dir(\vec{\alpha})$ is sampled.  To finally sample the items for a sequence, we use a multinomial distribution with the probability $p_i$ of an item $i$ given by the $i$th entry of $P_\sigma$.

For the number of items $N$ in a sequence, the literature proposes various options, most prominently, $N\in [50,200, 500]$ (see, e.g., \citep{bello2016neural}, \citep{kong2018new}, \citep{afshar2020state}, \citep{sur2022deep}, \citep{tu2023deep}). We choose $N=200$ as the sequence length because we argue that, for the practical application, we can take any sequence of length $N'$ and transform it to one or more sequences of length $200$. This is achieved either by waiting until $200$ items are accumulated or by splitting a larger sequence into subproblems of size 200.

\subsection{Extended Result Discussion}
\label{app:results}

Based on the results displayed in Tables \ref{tab:results_Uncorr}, \ref{tab:results_weakCorr} and \ref{tab:results_strongCorr} and Figure \ref{fig:decision-k3k7}, we can conclude that, in general, the difficulty of finding an optimal solution increases with the number of involved knapsacks for all problem specific approaches except CZL. The only exceptions to this trend are either very small ($\leq 0.12\%$) or due to the model's decision policy converging to a take all strategy. The two uninformed heuristics, random and take all, are able to decrease their relative gap to OR-Tools as the amount of correlation between weights and values increases. This makes sense as the influence of an uninformed decision decreases with more similar value-to-weight ratios of the items.
This trend is reversed for the problem-specific CZL heuristic as its main indication for a decision is the value-to-weight ratio of an item. While it is not obvious to see from the relative gap to OR-Tools, the performance of the heuristic gets closer to their worst case performance as the amount of correlation increases and, thus, the differences between the value-to-weight ratios decrease. 

On the uncorrelated dataset, the gap between the performance of the two reinforcement learning models, Kong and PostAlloc, is the smallest. The models perform significantly better than the online heuristics and the PostAlloc model achieves slightly better results being $1.07\%$ closer to the OR-Tools solution than Kong. The most significant difference in the gap to OR-Tools between Kong and PostAlloc is visible on the weakly correlated dataset. Both models converge to a take all policy once. Apart from that, the Kong model has not been able to extract a decision policy leading to significantly better results than the take all strategy. PostAlloc on the other hand is able to make better decisions for the datasets with $K \in [1,3,5]$, but it does not use the postpone action. With the strong correlation, Kong develops a take all policy for two of the datasets ($K \in [3,7]$), PostAlloc only for K=7. In the other three cases, PostAlloc outperforms the Kong model. For $K \in [3,5]$, the model performs even better then the greedy offline heuristic that achieved the best results on all other dataset configurations. In summary, the performance of our PostAlloc model is only $2.88\%$ below that of OR-Tools even though it does not require the knowledge about the full sequence in advance. That is $3.67\%$ closer than the model from Kong comes to OR-Tools. To calculate the reduction in the average product storage time, we first note that the PostAlloc model only used the additional postpone action on the uncorrelated datasets. This means that the results the model achieves on the other two datasets come with a reduction in average product storage time of $100\%$. On the uncorrelated dataset, the reduction is still very high with $96.1\%$ on average.